\documentclass[conference]{IEEEtran}
\IEEEoverridecommandlockouts

\usepackage{cite}
\usepackage{amsmath,amssymb,amsfonts}
\usepackage{algorithmic}
\usepackage{graphicx}
\usepackage{textcomp}
\usepackage{xcolor}
\usepackage{multirow}
\usepackage[hidelinks]{hyperref}
\def\BibTeX{{\rm B\kern-.05em{\sc i\kern-.025em b}\kern-.08em
    T\kern-.1667em\lower.7ex\hbox{E}\kern-.125emX}}
\begin{document}

\title{PortraVec: Image-Based Portrait Vectorization with Text-Guided Manipulation\\

\thanks{* Corresponding author}
\thanks{This work was supported by Project of Yuelushan Center for Industrial Innovation (Grant No.2025YCII0225), the National Natural Science Foundation of China (No.U25A20421), the National Key Research and Development Program of China (No.2025YFB3003601), Fundamental and Interdisciplinary Disciplines Breakthrough Plan of the Ministry of Education of China (No.JYB2025XDXM122)}
}

\author{
\IEEEauthorblockN{
Yiqi Liang, Ying Liu$^{*}$, Dandan Long, Ruihui Li
}
\IEEEauthorblockA{
\{yiqiliang, liu$\_$ying, ddlong, liruihui\}@hnu.edu.cn
}
}

\maketitle

\begin{abstract}
While portrait sketch generation is a special task in sketch synthesis, most existing methods are pixel-based, limiting their interpretability and editability.
With the rise of vector generation techniques, representing sketches using vector elements may provide more flexible manipulation.
However, due to the overlapping nature of vector graphics and the coarse detail modeling, existing vectorization methods struggle to capture facial integrity and fine-grained details, and lack semantic control.
To address these issues, we propose PortraVec, a framework for converting pixel-based portrait images into vector sketches with text control.
Specifically, we propose a two-stage image-guided generation module using Attention-aware Offset Sampling to capture face structure while correcting detail deviations, and a text-guided manipulation module based on Region-based Parameter Freezing to enable local semantic editing while maintaining global consistency.
Experiments show that PortraVec achieves superior structural consistency, visual fidelity, and semantic controllability compared to state-of-the-art methods.
\end{abstract}


\section{Introduction}
\label{sec:intro}
Portrait sketching is a compelling form of art and visual expression, widely used in fields such as art education, animation, and identity design~\cite{sun2023make, liang2025multi, lin2025inkspire}. 
Traditionally, creating and editing portrait sketches requires not only artistic skill, but also a precise grasp of geometry, proportion, and style. 

Recent techniques have led to a variety of methods that automatically generate portrait sketches from photographs~\cite{sun2025multi, yi2024generating, guo2025face}.
While effective at style rendering, these methods primarily operate at the pixel level, offering limited structural interpretability and editability—both essential for interactive and controllable applications. In contrast, vector graphics better reflect how artists conceptualize sketches by representing images as sequences of parametric curves (e.g., Bézier curves). This representation is resolution-independent, lightweight, and structurally editable, enabling intuitive manipulation, seamless scaling, and consistent visual quality. Consequently, vector-based generation provides a more suitable foundation for controllable and interactive sketch synthesis.
\begin{figure}[t]
    \centering
    \includegraphics[width=\columnwidth]{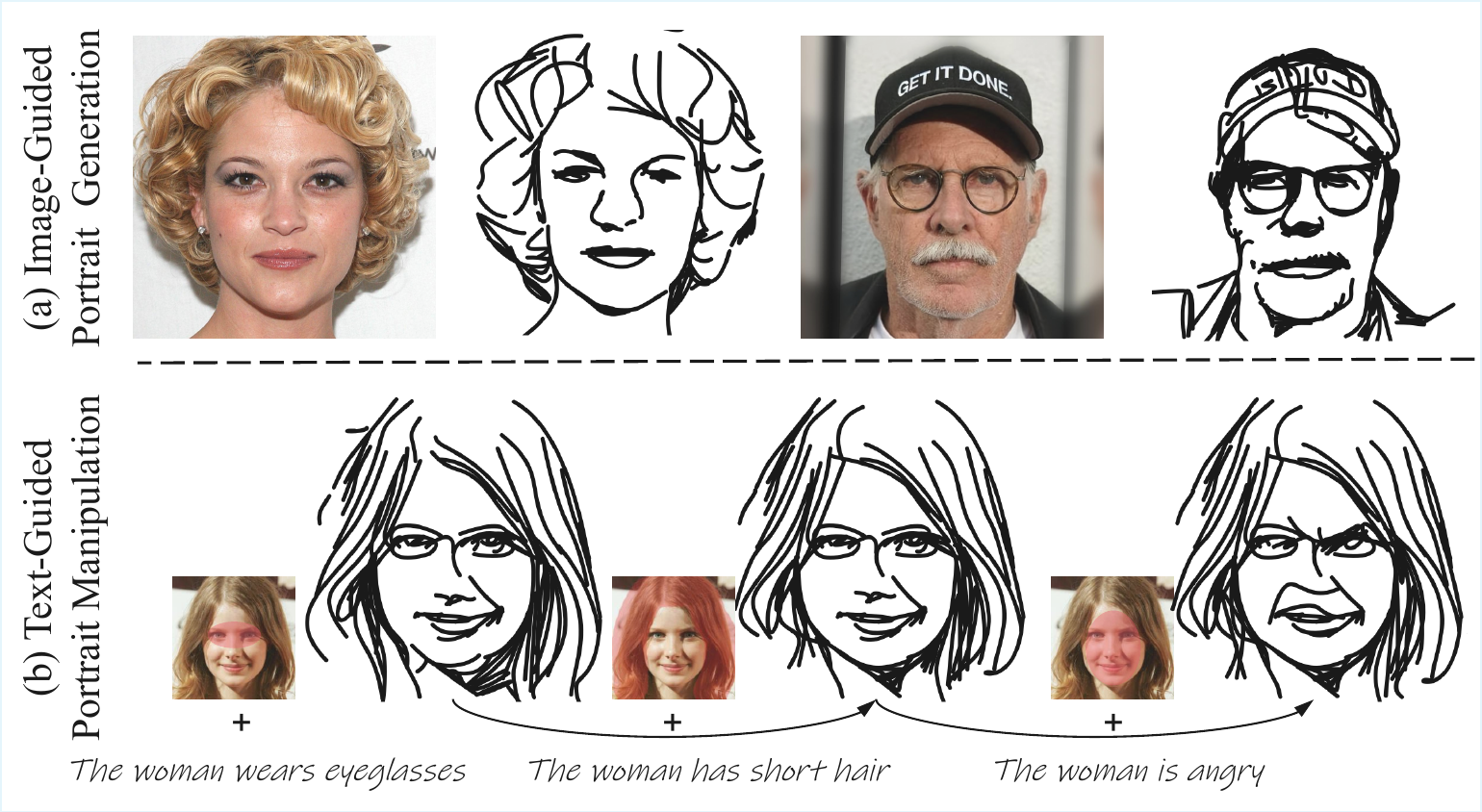} 
    \caption{PortraVec uses B\'{e}zier curves as sketch strokes to fully depict facial features at semantic, structural, and shading levels. (a) It preserves the global structure and key facial features across different abstraction levels (woman: 120 paths; man: 240 paths), ensuring consistent visual recognition; (b) With text guidance, PortraVec enables semantic manipulation of portrait sketches, achieving both fidelity to the input image and flexible, controllable editing.}
    \label{teaser}
\end{figure}

\begin{figure*}[t]
    \centering
    \includegraphics[width=\textwidth]{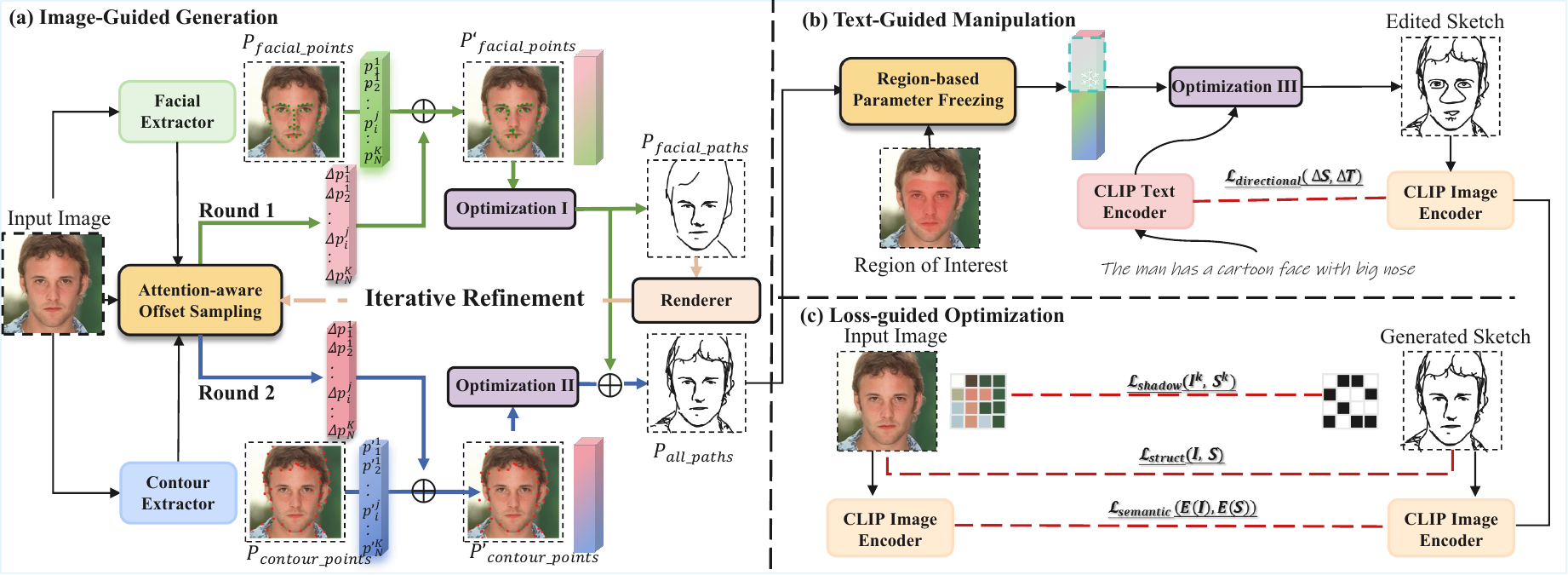} 
    \caption{Illustration of the PortraVec framework.
(a) The pipeline of the Image-Guided Generation module, which progressively converts an input face image into a vector portrait sketch through attention-aware sampling and optimization.
(b) The Text-Guided Manipulation module, enabling localized semantic editing of vector sketches via region-based parameter freezing while preserving global structure.
(c) An overview of the optimization losses, including semantic, structural, and shadow-aware terms, which jointly guide the vector sketch toward faithful appearance, coherent geometry, and visually consistent shading.}
    \label{network}
\end{figure*}
However, existing vector-based methods for portrait sketching struggle to preserve facial identity and semantic fidelity, especially when fine-grained details and geometric accuracy are required~\cite{vinker2022clipasso, frans2022clipdraw}. Moreover, semantic editing of portrait sketches remains challenging. Although multimodal models such as CLIP~\cite{radford2021learning} enable vision–language alignment, vector graphics (e.g., SVG) are still difficult for deep learning due to their hierarchical structures and diverse primitives. Recent works address this via hierarchical semantic alignment. For example, CLIPVG~\cite{song2023clipvg} enables text-guided vector manipulation via differentiable vector graphics without additional generative models, while SVGDreamer++~\cite{xing2025svgdreamer++} enhances editability and diversity through hierarchical vectorization and particle-based score distillation. Zhang et al.~\cite{zhang2023text} proposed a text-guided customization pipeline that preserves layer-wise properties of vectors. Despite these advances, portrait sketch manipulation remains difficult due to the complexity of facial structures.

To bridge these gaps, we propose PortraVec, a novel framework for vector-based portrait sketch generation and text-guided manipulation (Fig.~\ref{teaser}). PortraVec consists of two modules: image-guided generation and text-guided manipulation. In the image-guided module, facial landmarks and contours are used to initialize Bézier curve trajectories, followed by an Attention-aware Offset Sampling mechanism that shifts control points toward salient facial regions to enhance detail while preserving structure. A two-stage optimization strategy with semantic alignment, structural fidelity, and visual shadow losses ensures both expressiveness and geometric coherence.

For text-guided manipulation, PortraVec introduces a Region-based Parameter Freezing mechanism that confines optimization to semantically relevant facial regions, enabling localized edits (e.g., \textit{"make the eyebrows thicker"}) without global distortion. Additionally, PortraVec supports semantic interpolation between prompts and vector representations, allowing smooth transitions across styles or expressions. Extensive experiments demonstrate that PortraVec outperforms prior methods in visual quality and editability, advancing interactive and controllable portrait sketching in the vector domain.

In summary, our contributions are threefold: (1) We propose PortraVec, a novel framework for vector portrait sketch generation and manipulation, bridges the gap between geometric control and semantic guidance by leveraging both image and text inputs. (2) We introduce a progressive optimization strategy that enables fine-grained facial detail refinement and localized semantic editing without global distortion. (3) We demonstrate that PortraVec achieves state-of-the-art performance in portrait sketch generation and text-guided manipulation among existing pixel-based and vector-based methods.

\section{Methods}
The overall framework of PortraVec is shown in Figure~\ref{network}. Following optimization-based vector generation methods~\cite{frans2022clipdraw, vinker2022clipasso, vinker2023clipascene}, PortraVec iteratively optimizes vector parameters to generate sketches. It introduces two key modules—image-guided generation and text-guided manipulation—with tailored semantic, structural and shadow losses to guide optimization, ensuring the resulting sketches align with user intent while preserving facial identity and structural integrity.

 \subsection{Image-Guided Generation}\label{imguide}
As an optimization-based method, PortraVec first requires initialization of curve keypoint positions. Leveraging facial image characteristics, it employs face and contour extractors to obtain base curve points, then applies an attention-aware offset sampling mechanism to determine final curve paths for optimization. The curves are subsequently refined using semantic, structural, and shadow losses, producing sketches with both geometric consistency and visual expressiveness.

\textbf{Point Set Initialization. }\label{strokinit}
To ensure the integrity and structural accuracy of the portrait sketch, PortraVec uses a face extractor and a contour extractor to obtain the initial positions of key points of B\'ezier curve paths.
The face extractor applies a CNN~\cite{kazemi2014one} to detect the face region and extract facial landmarks $P_{facial\_points}$, which represent key facial components such as the eyes, nose, mouth, and eyebrows. The contour extractor first uses MaskGAN~\cite{lee2020maskgan} to segment the input image (Figure~\ref{shadow}), removes facial attribute masks to avoid redundancy, and retains non-facial elements such as skin, hair, hats, glasses, and neck. Canny edge detection~\cite{ding2001canny} is then applied to the mask to extract contour keypoints $P_{contour\_points}$.

\textbf{Attention-aware Offset Sampling. }
\begin{figure}[t]
    \centering
    \includegraphics[width=\columnwidth]{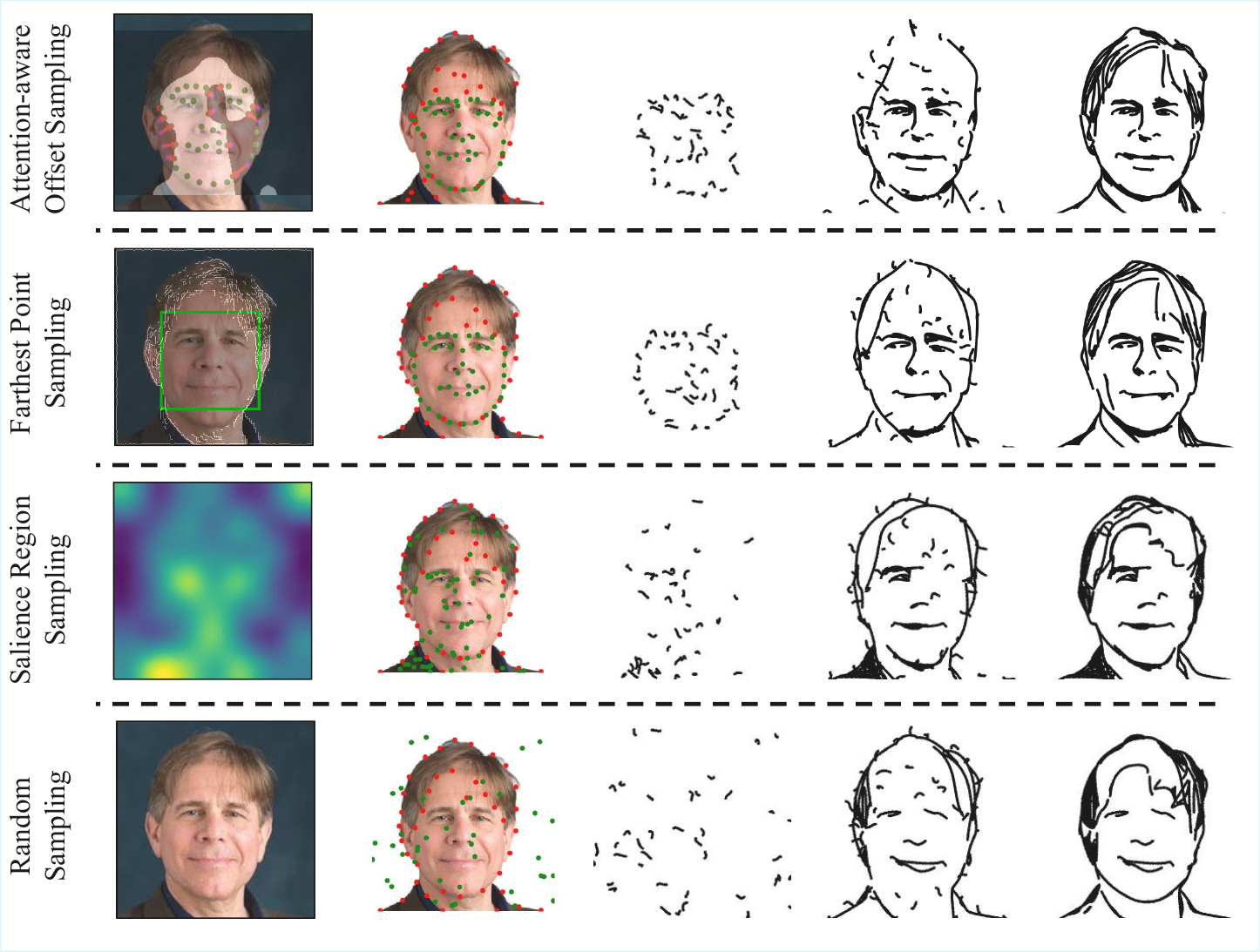} 
\caption{Comparison of different curve initialization sampling strategies.
From top to bottom:
(1) our attention-guided sampling, which shifts uniformly distributed points toward high-attention regions;
(2) Farthest Point Sampling (FPS) used in MROSS~\cite{liang2025mross}, which samples points along facial contours;
(3) Salient Region Sampling in CLIPasso~\cite{vinker2022clipasso}, based on image saliency;
and (4) random sampling as a baseline.
All sampled points serve as initial control points for curve generation, and all methods follow the same two-stage optimization for fair comparison.
}
    \label{sample}
\end{figure}
In vector sketch generation, directly sampling locations from face--derived and contour-derived point sets is insufficient, as these points fail to capture key facial semantic structures, severely affecting structural and visual quality. Existing methods, such as CLIPasso~\cite{vinker2022clipasso} and MROSS~\cite{liang2025mross}, rely on salient-region or uniform FPS sampling, but remain inadequate for portrait sketches that require precise facial structure and fine details (Fig.~\ref{sample}).

\begin{figure}[t]
    \centering
    \includegraphics[width=\columnwidth]{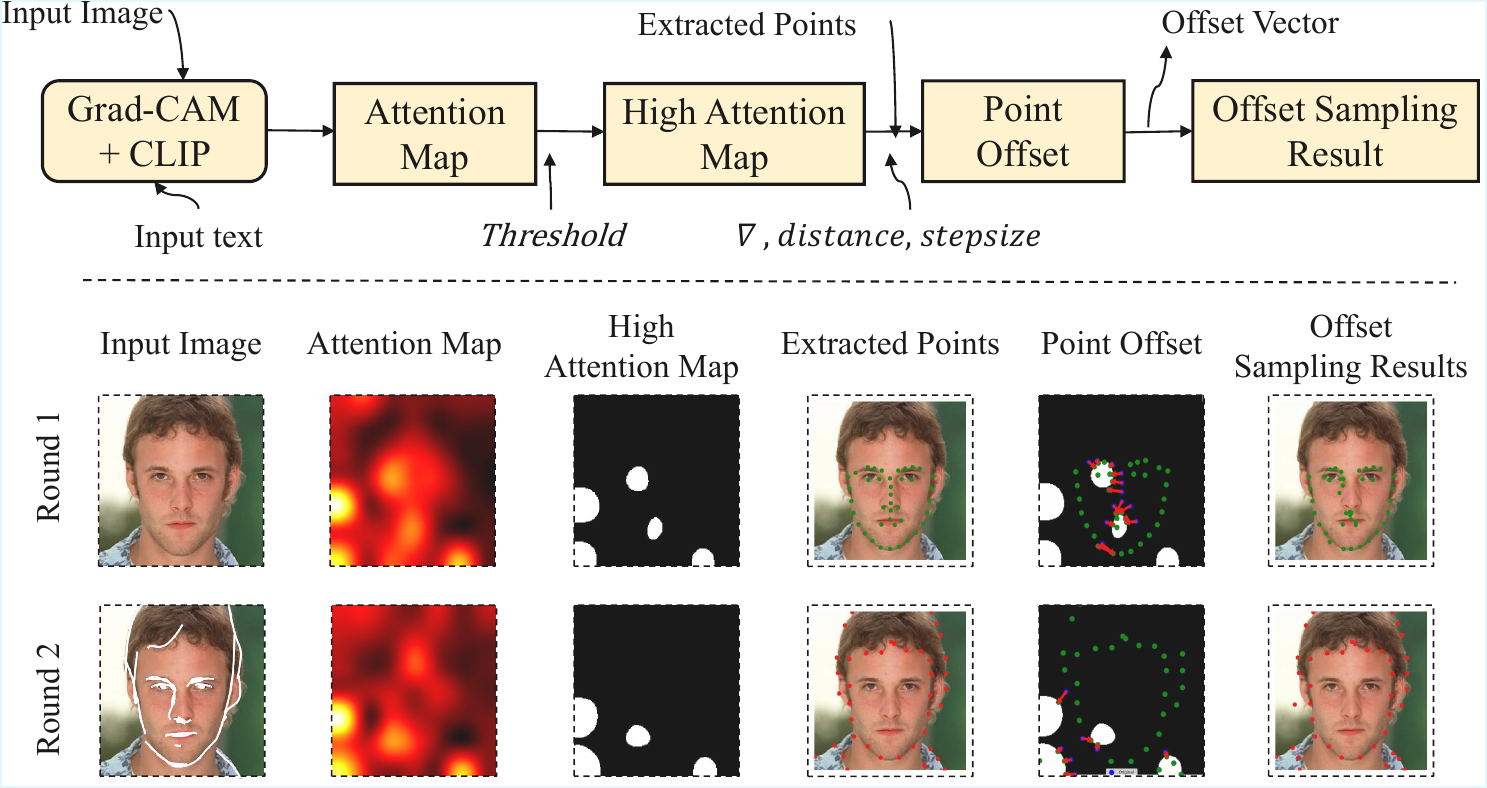} 
    \caption{The pipeline and two-round visualized examples of the proposed Attention-aware Offset Sampling mechanism.}
    \label{attn_sample}
\end{figure}

To address this, PortraVec further introduces an \emph{Attention-aware Offset Sampling} mechanism that integrates semantic saliency with geometric control (Fig.~\ref{attn_sample}).
Given an input image and a text prompt, we first compute a \emph{text-conditioned attention map} using Grad-CAM~\cite{selvaraju2017grad} on the CLIP image encoder. The cosine similarity between image and text embeddings serves as the supervision signal, producing an attention map highlighting semantically important facial regions. We extract high-attention regions by thresholding the map. Each point in the set is refined by applying an offset toward nearby high-attention areas. The offset direction follows the local attention gradient, while the step size is scaled by the corresponding attention magnitude, encouraging points to concentrate on perceptually salient structures. This process yields an attention-guided first-round point set $P'_{\mathrm{facial\_points}}$.
Those points are fitted with B\'ezier curves to generate facial stroke paths and rendered into the first-round vector sketch. To further enhance contour details, we utilize a second-round offset strategy. Specifically, we compute a residual map between the input image and the first-round rendered sketch, which highlights underrepresented or missing regions. The residual map is processed by the same Grad-CAM pipeline to obtain a second attention map, guiding the offset of contour points, denote as $P'_{\mathrm{contour\_points}}$.
Finally, the second-round contour paths are merged with the first-round facial paths to form the complete path set $P'_{\mathrm{all\_paths}}$. 
Through this two-round sampling process, ProtraVec produces sketches that are both structurally coherent and detail-rich, as demonstrated in Fig.~\ref{attn_sample}.

\textbf{Image-Guided Optimization. }
PortraVec employs a two-round image-guided optimization: the first round refines facial points for a rough sketch, and the second refines contour points to add details, with all parameters building on the previous round. Typically, we run 500 iterations for the first round and 200 for the second, achieving convergence and a stable portrait sketch in about 5 minutes.

 \subsection{Text-Guided Manipulation}\label{textguide}
The text-guided manipulation module modifies global or specific facial regions (e.g., \textit{"make the corners of the mouth smile"}) without affecting unrelated areas. To enable precise editing, PortraVec employs a Region-based Parameter Freezing mechanism, which freezes irrelevant curves and performs text-driven optimization only on the unfrozen parameters for localized adjustments.

\textbf{Region-based Parameter Freezing.}
Sketch curves are represented parametrically using Bézier control points and stroke attributes. For localized semantic editing, users specify a Region of Interest (ROI) in image space. We define a curve-level mapping
$f: \mathrm{ROI} \rightarrow \mathcal{C}_R \subseteq \mathcal{C}$,
where a curve $c_i \in \mathcal{C}$ is included in $\mathcal{C}_R$ if it intersects the ROI. Once selected, all control points of $c_i$ are treated as region-relevant, regardless of their individual positions.

During optimization, only Bézier control points of curves in $\mathcal{C}_R$ are updated, while curves outside the ROI remain frozen. Stroke-level attributes (e.g., width and color) are fixed to maintain global style consistency. When multiple ROIs overlap, their corresponding curve sets are merged by union. This region-based parameter masking enables high-fidelity, semantically controllable, and expressive portrait sketch editing.

\textbf{Text-Guided Optimization. }
PortraVec uses a single-round optimization mechanism in text-guided manipulation module. 
Based on the input sketch and the ROI, PortraVec optimizes the unfrozen curve parameters until the optimization error no longer changes significantly, with about 500 iterations per round (taking 3 minute).

\subsection{Loss Function}
Since each round of optimization has different purposes, different loss functions are used for gradient calculation. In the image-guided generation module, the first round of optimization is guided by $\mathcal{L}_{\text{semantic}}$.
In addition to $\mathcal{L}_{\text{semantic}}$, the second round  of optimization is also guided by $\mathcal{L}_{\text{structure}}$ and $\mathcal{L}_{\text{show}}$. 
In the text-guided manipulation module, besides the preservation of semantics, structure and shadows, $\mathcal{L}_{\text{directional}}$ is supplemented to conform to the text content.

\textbf{CLIP-based Semantic Loss. }Since sketches consist only of curves, pixel-level metrics are inadequate for measuring similarity to images. Following CLIPasso~\cite{vinker2022clipasso}, we use multi-layer features from CLIP (ResNet-101) to guide generation. The loss between the image $\mathcal{I}$ and sketch $\mathcal{S}$ is defined as:

\begin{equation}\label{Loss_clip}
          \begin{aligned}
    \mathcal{L}_{\text{semantic}} &= \text{\textit{dist}}(CLIP(\mathcal{I}), CLIP(\mathcal{S})) \\
    &\quad + L_2(CLIP_{l_{2,3,4}}(\mathcal{I}), CLIP_{l_{2,3,4}}(\mathcal{S}))
\end{aligned}
\end{equation}
where $\text{\textit{dist}}(\cdot,\cdot)$ denotes cosine distance between global CLIP embeddings, and $CLIP_{l_{2,3,4}}(\cdot)$ are intermediate-layer features.

\textbf{VGG-based Structure Loss. }To make the sketch more reflective of the contours of the input image, we compute the Learned Perceptual Image Patch Similarity (LPIPS)~\cite{zhang2018unreasonable} between the sketch and the mask image based on VGG16~\cite{simonyan2014very}. 
$I_{mask}$ is the masked image obtained from input image $I$.
\begin{equation}\label{Loss_vgg}
            \mathcal{L}_{structure} = {LPIPS(I_{mask},S)}
\end{equation}

\begin{figure}[t]
    \centering
    \includegraphics[width=\columnwidth]{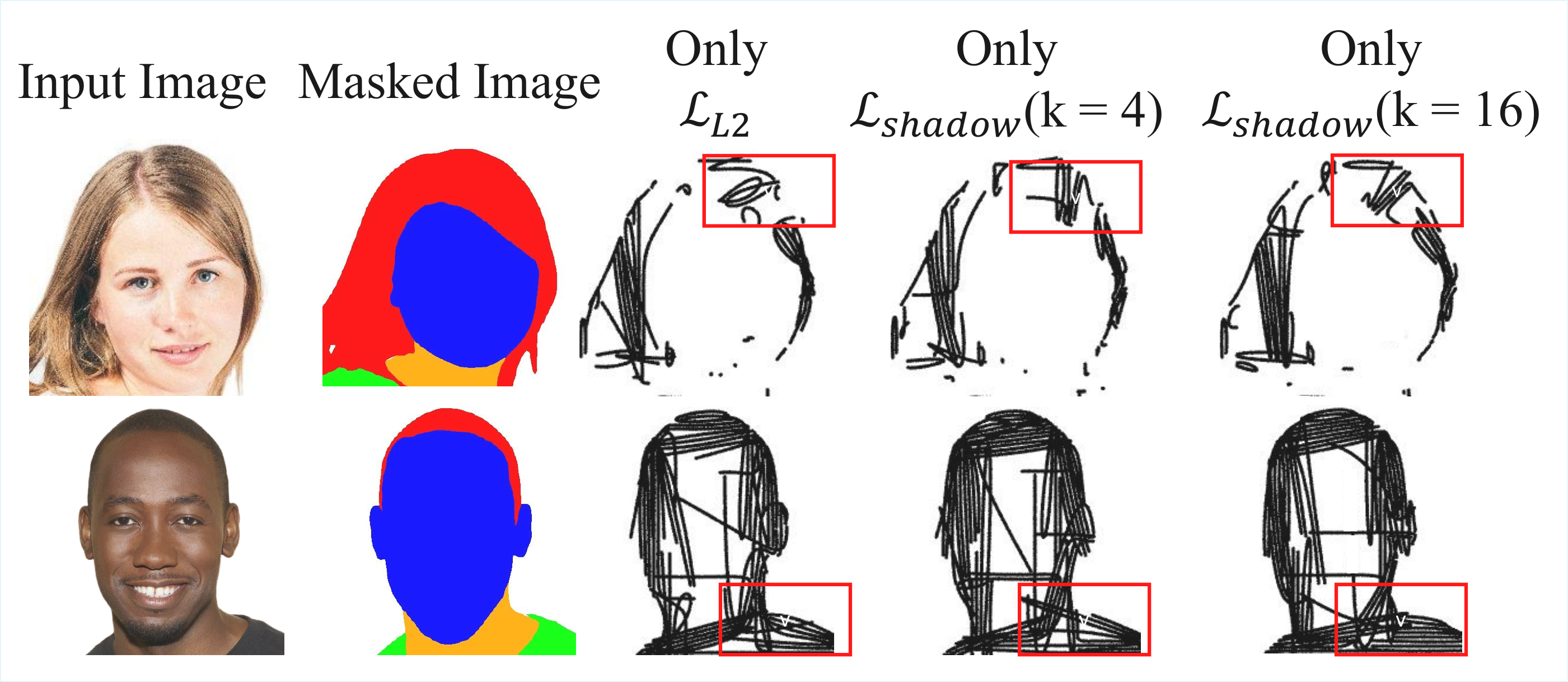} 
        \caption{Results for different k settings. Using only a global $\mathcal{L}_{L2}$ over-smooths shadow regions. When $k$ is small, cropped blocks average shadow and non-shadow areas, resulting in coarse shading. Increasing $k$ enforces localized shadow constraints; however, overly large $k$ may introduce fragmented and noisy strokes.
Empirically, $k=16$ is sufficient for $224 \times 224$ images.}
    \label{shadow}
\end{figure}
\textbf{Crop-based Shadow Loss. }
Inspired by Clipasso \cite{vinker2022clipasso}, we observe that the L2 loss mainly encourages pixel-wise filling, which tends to blur fine tonal transitions in portrait sketches. 
To better preserve shadow structures while maintaining sketch clarity, we propose a crop-based shadow loss that combines global and local L2 constraints:

\begin{equation}\label{Loss_crop}
\mathcal{L}_{shadow}
=
\left\|\mathcal{I}-\mathcal{S}\right\|_2^2
+
\frac{1}{k}
\sum_{i=1}^{k}
\left\|
\mathcal{I}_i - \mathcal{S}_i
\right\|_2^2 ,
\end{equation}

where $\mathcal{I}_i$ and $\mathcal{S}_i$ denote the $i$-th cropped block from the input image and sketch respectively. $k$ determines the spatial granularity of shadow supervision (effects shown in Figure~\ref{shadow}).


\textbf{CLIP-based Directional Loss. }Due to the sparsity of curve representations, we adopt a directional loss from StyleGAN-NADA~\cite{patashnik2021styleclip}. We compute the semantic direction between the source and target sketches in CLIP embedding space, and during optimization, fine-tune the unfrozen path parameters to align the generated sketch with the text-implied semantic shift. The loss is formally defined as:

\begin{equation}\label{Loss_text}
    \mathcal{L}_{\mathrm{directional}}=1-\cos\left(\Delta S,\Delta T\right)
\end{equation}

where $\Delta S=\mathrm{CLIP}(S_{\mathrm{edit}})-\mathrm{CLIP}(S_{\mathrm{generate}})$ and 
$\Delta T=\mathrm{CLIP}(T_{\mathrm{edit}})-\mathrm{CLIP}(T)$ denote the sketch and text editing directions, respectively.
$S_{\mathrm{edit}}$ and $T_{\mathrm{edit}}$ represent the edited sketch and target text prompt, while $T$ is set to \textit{``man''} or \textit{``woman''} by default.


\section{Experiments} 
\subsection{Basic Stroke and Renderer Settings}
\begin{table}[t]
\caption{Quantitative comparison of image-guided generation. 
$\downarrow$ means the lower the better while $\uparrow$ means the opposite. 
$/$ before and after are controls with 30 and 120 paths.}
\label{tab:generate}
\centering
\setlength{\tabcolsep}{4pt}
\renewcommand{\arraystretch}{1.2}
\begin{tabular}{c|c|c|c|c}
\hline
Method & LPIPS $\downarrow$ & SSIM $\uparrow$ & Acc. (\%) $\uparrow$ & Pref. (\%) $\uparrow$ \\
\hline
Virtual Sketching 
& 0.648 
& 0.567 
& 58.0 
& 0 \\
\hline
Clipasso 
& 0.611 / 0.572 
& 0.565 / 0.583 
& 76.0 / 82.0 
& 14.1 / 22.5 \\
\hline
SVGDreamer  
& 0.598 / 0.555 
& 0.575 / 0.609 
& 84.0 / 74.0 
& 7.0 / 19.4 \\
\hline
SwiftSketch  
& 0.586 / 0.540 
& 0.570 / 0.611 
& 79.0 / 71.0 
& 1.5 / 8.9 \\
\hline
PortraVec 
& \textbf{0.582} / \textbf{0.539} 
& \textbf{0.587} / \textbf{0.629} 
& \textbf{91.0} / \textbf{96.0} 
& \textbf{77.2} / \textbf{50.0} \\
\hline
\end{tabular}
\end{table}

\begin{table}[t]
\caption{Quantitative comparison of text-guided manipulation. 
$\uparrow$ means better effect.}
\label{tab:manipulate}
\centering
\setlength{\tabcolsep}{6pt}
\renewcommand{\arraystretch}{1.2}
\begin{tabular}{c|c|c|c|c}
\hline
Setting & Metric & CLIPVG & VectorFusion & Ours \\
\hline
\multirow{2}{*}{Full Image}
& Acc. (\%) $\uparrow$  & 76.4 & 43.8 & \textbf{79.8} \\
\cline{2-5}
& Pref. (\%) $\uparrow$ & 33.7 & 15.0 & \textbf{51.1} \\
\hline
\multirow{2}{*}{Region of Interest}
& Acc. (\%) $\uparrow$  & 63.1 & 70.6 & \textbf{80.8} \\
\cline{2-5}
& Pref. (\%) $\uparrow$ & 17.0 & 34.0 & \textbf{47.0} \\
\hline
\end{tabular}
\end{table}
Each sketch is parameterized as a set of cubic Bézier curves defined by four control points, providing sufficient expressive power for smooth stroke representation while remaining amenable to continuous optimization. All strokes share a fixed width of 0.5 and are rendered in uniform black, forming unified sketch effects. All vector sketches are represented and optimized using a differentiable renderer~\cite{li2020differentiable}, enabling pixel-level losses to back-propagate gradients to vector parameters.

\subsection{Dataset and Baseline Models} 
Experiments are conducted on CelebA-HQ~\cite{karras2017progressive}, a 30,000-image high-quality subset of CelebA~\cite{liu2015deep} at 1024×1024. For fair comparison, we randomly sample and resize images to 224×224 to match model inputs (e.g., CLIP, VGG). PortraVec is compared with Virtual Sketching~\cite{mo2021general}, CLIPasso~\cite{vinker2022clipasso}, SVGDreamer~\cite{xing2024svgdreamer}, and SwiftSketch~\cite{arar2025swiftsketch} for image-guided generation, and with VectorFusion~\cite{jain2023vectorfusion} and CLIPVG~\cite{song2023clipvg} for text-guided manipulation. Additional results on occluded and profile faces are in the supplementary material.

\begin{figure*}[t]
    \centering
    \includegraphics[width=\textwidth]{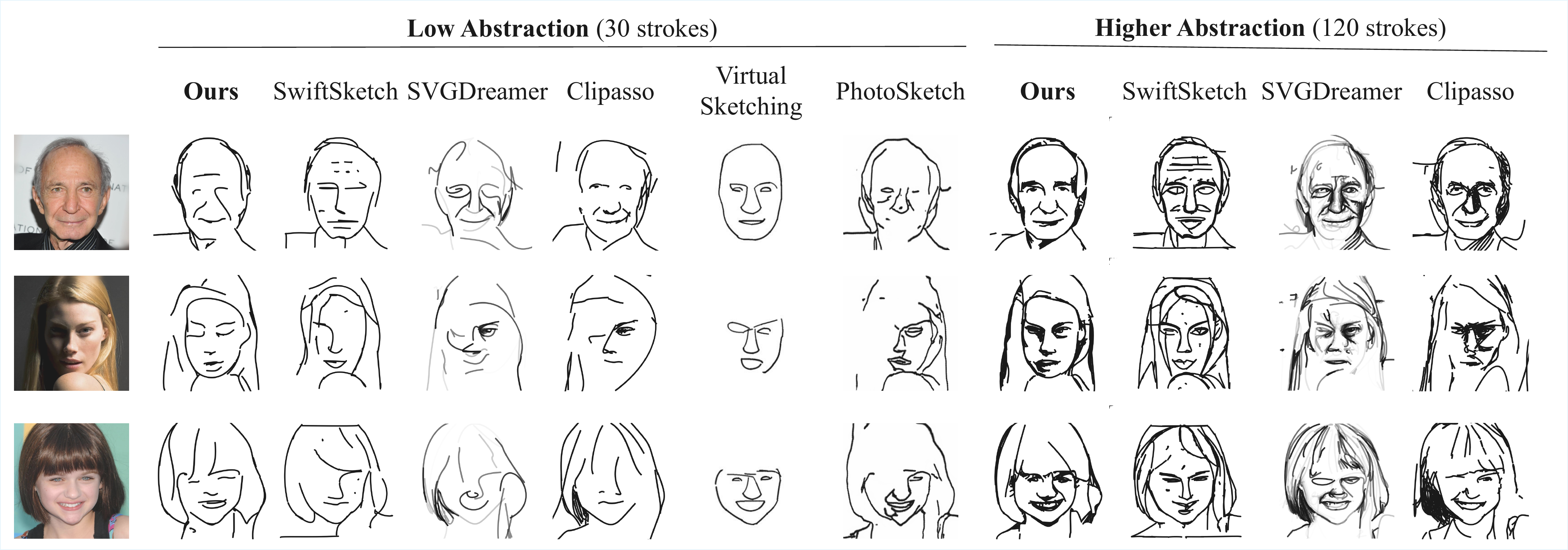} 
    \caption{Comparison of portrait results with state-of-the-art methods. To ensure a fair comparison, sketches are generated at two abstraction levels—low (30 vector paths) and high (120 vector paths)—due to the limited visual control in Virtual Sketching.}
    \label{compare}
\end{figure*}
\subsection{Evaluation Metrics}
We evaluate our method on structural consistency, diversity, accuracy, and visual preference~\cite{li2020lightweight}. LPIPS~\cite{zhang2018unreasonable} and SSIM~\cite{wang2004image} measure diversity and structural consistency, while accuracy and visual preference are assessed via a user study. We randomly select 50 images from CelebA-HQ~\cite{karras2017progressive}, generating 30-path and 120-path sketches per image. For fairness, Virtual Sketching~\cite{mo2021general} is included in the 30-path group, and text edits are applied to the 120-path group for comparison.

\subsection{Comparison of Image-Guided Generation}

\textbf{Quantitative Comparison.}
We evaluate performance using LPIPS, SSIM, as well as visual accuracy and preference. Table~\ref{tab:generate} shows that PortraVec outperforms state-of-the-art methods across all metrics, demonstrating its ability to generate complete, diverse, accurate, and visually appealing sketches.

\begin{figure}[t]
    \centering
    \includegraphics[width=\columnwidth]{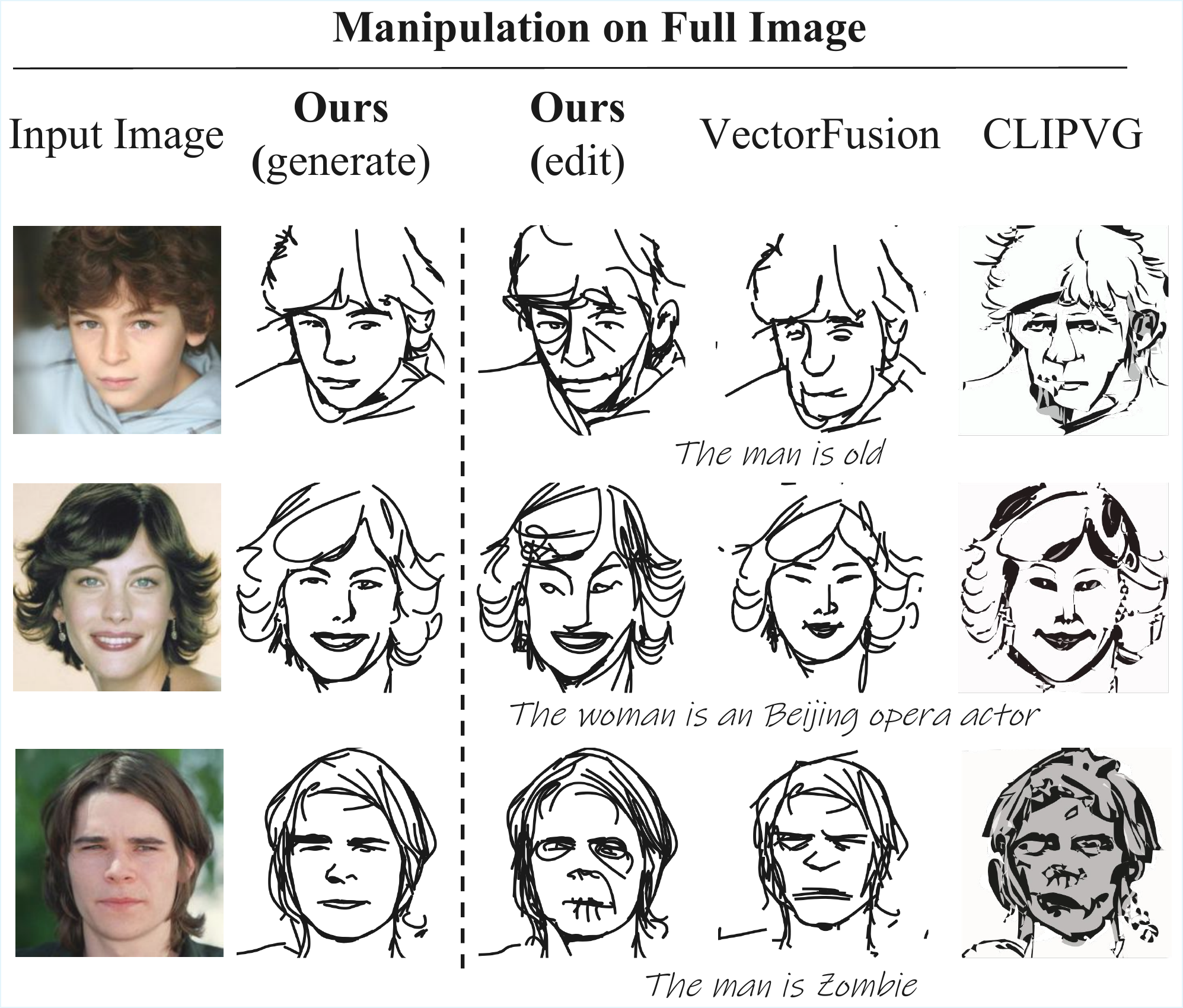} 
    \caption{Comparison of portrait sketches with text-guided manipulation on full images. PortraVec delivers more coherent edits than other works.}
    \label{full_edit}
\end{figure}

\begin{figure}[t]
    \centering
    \includegraphics[width=\columnwidth]{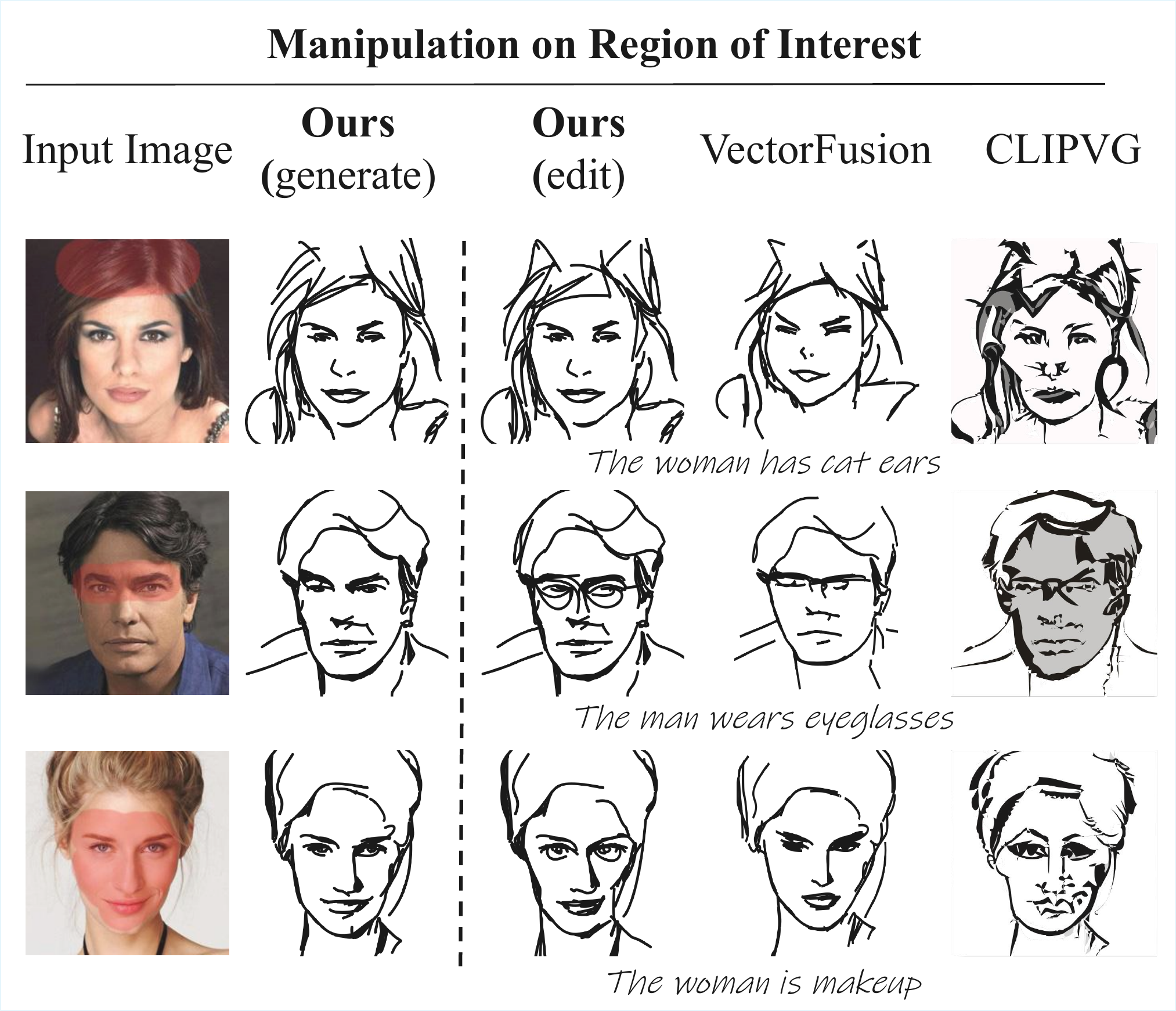} 
    \caption{Comparison of portrait sketches with text-guided manipulation on regions of interest. PortraVec delivers more accurate edits than other works.}
    \label{roi_edit}
\end{figure}
\textbf{Qualitative Comparison. }
In Figure~\ref{compare}, we compare PortraVec with five state-of-the-art vector sketch methods under two abstraction levels. Virtual Sketching struggles with portraits, producing incorrect topology and missing details. CLIPasso~\cite{vinker2022clipasso}, SVGDreamer~\cite{xing2024svgdreamer}, and SwiftSketch~\cite{arar2025swiftsketch} better manage abstraction, but low-path outputs miss facial features and high-path outputs are cluttered. PortraVec generates coherent and structurally accurate portraits at both levels.

\subsection{Comparison of Text-Guided Manipulation}
\textbf{Quantitative Comparison. }
We conducted a user study with randomly selected images with randomly selected descriptions. 
The results are shown in Table~\ref{tab:manipulate}.
Compared with CLIPVG~\cite{song2023clipvg} and VectorFusion~\cite{jain2023vectorfusion}, ProtraVec achieves better accuracy, and visual preference. 

\textbf{Qualitative Comparison. }
\begin{figure}[t]
    \centering
    \includegraphics[width=\columnwidth]{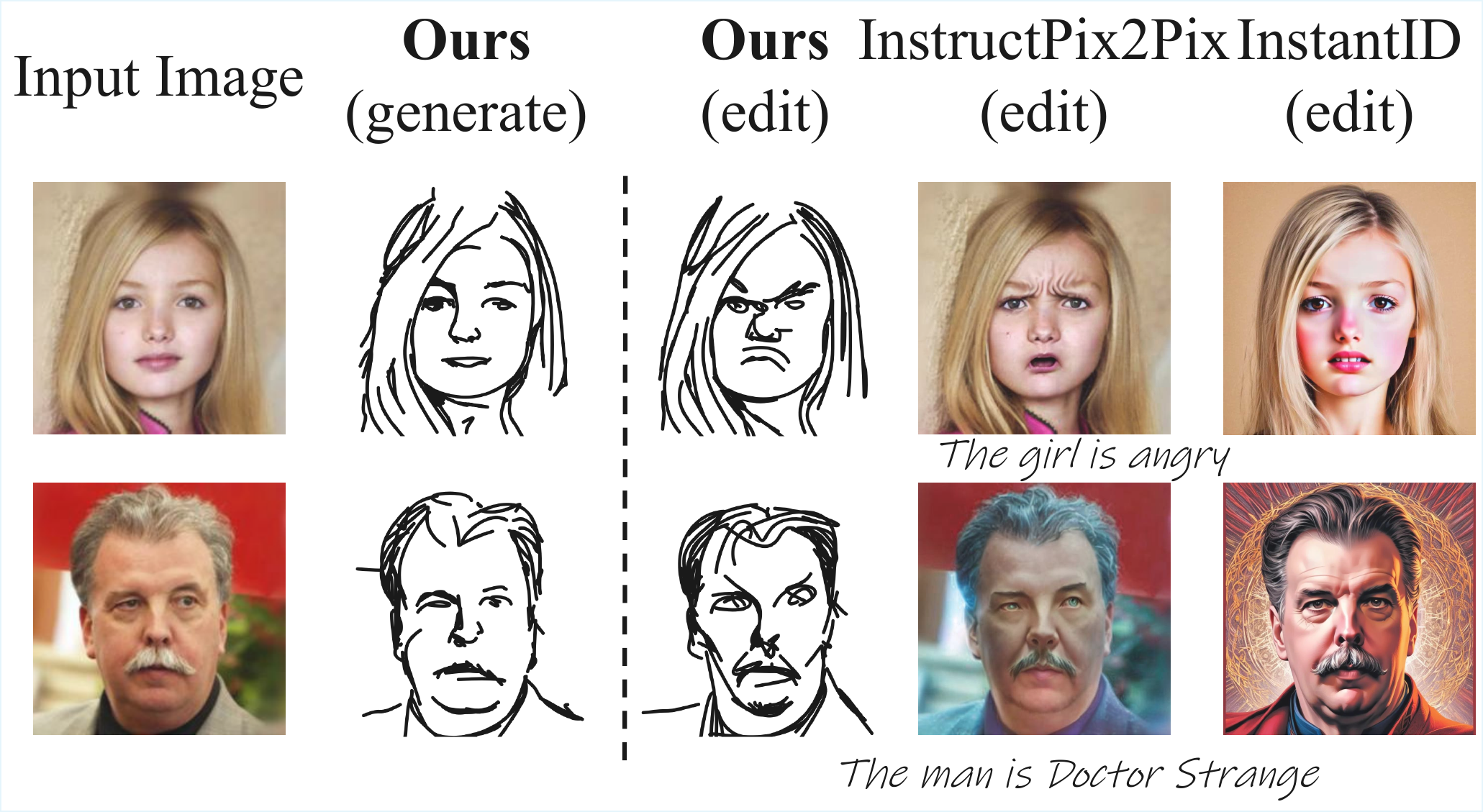} 
    \caption{Comparison between ProtraVec and pixel-based methods. ProtraVec allows for more drastic changes in facial geometry and emotional expression.}
    \label{pixel_compare}
\end{figure}
Figures~\ref{full_edit} and~\ref{roi_edit} show editing results on full images and ROIs under various text prompts. CLIPVG~\cite{song2023clipvg} often distorts or introduces artifacts, and VectorFusion~\cite{jain2023vectorfusion} always alters irrelevant areas. PortraVec delivers accurate, clean edits that preserve semantic relevance and identity. Figure~\ref{pixel_compare} compares PortraVec with pixel-based methods InstructPix2Pix~\cite{brooks2023instructpix2pix} and InstantID~\cite{wang2024instantid}, showing more pronounced expressions while maintaining identity. 

\section{Conclusion}
PortraVec is a novel optimization-based framework that unifies image-based sketch generation and text-guided manipulation into a single pipeline. By introducing an attention-aware offset sampling mechanism and a region-based parameter freezing mechanism, it achieves high semantic precision, structural fidelity, and local controllability in portrait sketch generation. Supported by designed loss functions and multimodal guidance, PortraVec produces sketches with enhanced geometric consistency and artistic expression. Extensive experiments demonstrate its superiority in generating expressive, editable, and identity-preserving portrait sketches.

\bibliographystyle{IEEEbib}
\bibliography{icme2026references}

\end{document}